\documentclass{article}

\usepackage{microtype}
\usepackage{graphicx}
\usepackage{subcaption}
\usepackage{booktabs}
\usepackage{flafter}
\usepackage{hyperref}
\usepackage{amsmath,amssymb}
\usepackage{mathtools}
\usepackage{amsthm}
\usepackage[capitalize,noabbrev]{cleveref}

\theoremstyle{plain}

\theoremstyle{definition}

\theoremstyle{remark}

\usepackage{algorithm}
\usepackage{algorithmic}

\usepackage[accepted]{icml2026}
\usepackage{xcolor}

\makeatletter
\renewcommand{\Notice@String}{Accepted for a spotlight at ICML 2026 Workshop on Generative and Agentic AI for Biology (GenBio) and accepted as a poster at ICML 2026 Workshop on Decision-Making from Offline Datasets to Online Adaptation: Black-Box Optimization to Reinforcement Learning (DEMO).}
\makeatother

\icmltitlerunning{Learning Clinical-Trial Strategy: Offline Policy Training for Decision Agents}

\begin{document}

\twocolumn[
\icmltitle{Learning Clinical-Trial Strategy: Offline Policy Training for Decision Agents}

\begin{icmlauthorlist}
\icmlauthor{William Bolton}{aff1}
\icmlauthor{Philip Torr}{aff1}
\end{icmlauthorlist}

\icmlaffiliation{aff1}{Department of Engineering Science, University of Oxford, Oxford, United Kingdom}
\icmlcorrespondingauthor{William Bolton}{will.bolton@eng.ox.ac.uk}
\icmlkeywords{Agentic AI, decision agents, LLM, sequential decision-making, offline policy learning, clinical-trials, strategy}

\vskip 0.3in
]

\printAffiliationsAndNotice{}

\begin{abstract}
Clinical development is sequential decision-making under uncertainty, where a sponsor must plan a portfolio of experiments from heterogeneous evidence. We study this setting by framing oncology clinical development as an offline decision-making problem in which an agent predicts the next six-month trial portfolio of an oncology drug program from information available at the decision date. To support this, we construct a temporal dataset that combines 31.7k heterogeneous public data records, including trial registries, regulatory reviews, sponsor filings, utilization data, and epidemiology, into 881 offline decision episodes across 45 historical programs. We compare four offline objectives: behavioral cloning, reward-weighted behavioral cloning, learned-reward training, and value-based implicit Q-learning against four frontier LLM agents that share a common date-gated retrieval scaffold across held-out drug, sponsor, drug-class, and temporal splits. Models trained offline outperform the non-fine-tuned baselines, particularly in the post-August 2025 contamination-clean holdout. Reward-weighted behavioral cloning performs the best, obtaining 46.2\% indication F1 and 14.2\% strict F1 against 25.0\% and 2.1\%, respectively, for the best-performing tool agent on each metric. These results suggest that structured offline learning can teach agents to plan clinical experiments.
\end{abstract}

\section{Introduction}
\label{sec:intro}

Clinical development is a sequential decision-making problem under uncertainty. At each point in a drug program, a sponsor must decide what to plan next, whether to expand into a new indication, move to a later phase, test a combination, change geography, collect real-world evidence, or stop investing. These decisions are made with imperfect information. Sponsors combine prior trials, regulatory signals, competitor activity, epidemiology, and commercial context, but future clinical, regulatory, and market outcomes remain unknown. Furthermore, the consequences of decisions are delayed, since a trial must run for its outcome to be observed, and it may only affect an approval, revenue, or patients years in the future.

Most machine-learning work on clinical trials asks whether a predefined trial will succeed or investigates problems such as whether a patient matches a trial \citep{fu2022hint,jin2024matching,chen2025trialbench}. We instead study the upstream strategy. Given the state of a drug program and its external environment at time $t$, what trials should be launched next? 

We instantiate this problem in oncology by constructing a temporal dataset that turns heterogeneous public evidence into structured states, trial-portfolio actions, and delayed outcome signals for offline sequential decision-making. Methodologically, this work connects to offline reinforcement learning and sequence modeling from logged trajectories, where policies are learned from fixed datasets rather than through online interaction \citep{levine2020offline,prudencio2023survey,chen2021decision,janner2021offline}. Our reward-weighted objective is closest to advantage-weighted behavioral cloning, where imitation learning is reweighted by return or advantage estimates \citep{peng2019awr,nair2020awac,kostrikov2021offline}. It also relates to LLM-agent work in which models use tools to support scientific experimentation, and simulated clinical reasoning \citep{huang2025biomni,schmidgall2024agentclinic}. The key difference is that our actions are timestamped trial portfolios, the evidence is date-gated, and rewards come from external outcomes observed after the decision. This work is a first step toward models that learn from the history of clinical development to support better trial strategy under uncertainty.

\paragraph{Contributions.}
\begin{itemize}
    \item We formulate clinical-trial strategy as an offline training problem over
    structured trial portfolios.
    \item We construct a temporal oncology dataset of 881 decision episodes across 45 drug programs, assembled from 31.7k  public records spanning trial registries, FDA and EMA reviews, SEC filings, CMS utilization, SEER epidemiology, and literature, yielding 27k structured fields to predict across 1,956 launch actions.
    \item We compare four offline training objectives against four frontier LLM agents on a shared date-gated retrieval scaffold, across drug, sponsor, drug-class, and temporal splits.
\end{itemize}

\section{Methods}
\label{sec:method}

\subsection{Problem formulation}
\label{sec:method-formulation}

Clinical-development strategy is formulated as an offline decision-making problem over historical drug-program windows. For each drug program $p$ and each six-month window $t$, the model observes a state $s_{p,t}$ containing only information available before $t$, including completed and ongoing trials, prior approvals, the competitive landscape, epidemiology, and market positioning. It then predicts an action $a_{p,t}$, defined as the set of trials launched by the sponsor during that window, with each trial represented as a 14-field structured object. Each program-window pair is also assigned an outcome-derived reward $R_{p,t}$, which summarizes downstream real-world signals associated with those decisions. 

\subsection{Dataset construction}
\label{sec:method-data}

We assembled a novel temporal dataset on oncology drug development
programs by curating public information. Collection was seeded by per-drug configuration files containing canonical drug names, synonyms, sponsor identifiers, regulatory keys, and source-specific queries. This was necessary because across different sources the same drug can appear under different brand names, international non-proprietary names, development codes, and sponsor aliases.

The episode state, action, and reward are constructed from five primary public-source streams. ClinicalTrials.gov provides trials, eligibility, outcomes, and trial timing, with each registered trial identified by a unique \texttt{NCT} accession number. Drugs@FDA provides review PDFs that our approval linker mines for those NCT identifiers and trial acronyms. SEC EDGAR provides filings parsed for quarterly revenue. CMS Part D provides beneficiary counts, and SEER provides US cancer epidemiology. Raw fetches are stored immutably with content-hash addressing and full provenance, then normalized into trial records, document records, quarterly revenue entries, and per-trial competitive snapshots evaluated at each trial start date.

FDA approvals are linked back to supporting ClinicalTrials.gov trials using NCT identifier matches in review documents, trial-acronym matching, and sponsor-scoped fallback rules. This allows approval credit to be assigned to the decision window in which the linked pivotal trial was launched, allowing delayed approvals to be credited to the earlier launch decision. The resulting offline dataset contains 881 historical $(s,a,R)$ triples across 45 drug programs.

The 881 windows are partitioned into eight held-out evaluation splits that test different generalization axes. Five single-drug holdouts (tagrisso, gleevec, imbruvica, keytruda, opdivo) hold out every window of one drug program to test transfer to a fully unseen drug with rich training history. A sponsor-blind split holds out all four AstraZeneca programs (tagrisso, lynparza, imfinzi, calquence) to test transfer between sponsors. The drug-class split holds out all PD-1 and PD-L1 checkpoint inhibitors to test transfer across pharmacology. Finally, two temporal splits hold out windows with a decision date after a model's knowledge cutoffs. A post-September 2024 split aligned with the Qwen-2.5 cutoff, and a post-August 2025 split aligned with the latest cutoff for GPT-5.4 and Opus 4.5, to provide a prospective assessment of model performance that cannot be contaminated by training data. Each split's remaining windows form its own training set, and together the eight test sets total 432 held-out windows.

\subsection{Action representation}
\label{sec:method-action}

Every predicted and ground-truth trial conforms to a JSON Schema
enforced at decode time. The schema fixes 14 required fields per trial, including the indication, phase, strategy, study design, enrollment bucket, comparator, region, and endpoint fields, with categorical vocabularies enumerated where appropriate. Structured decoding ensures that locally served models emit the same field set, making per-dimension F1 directly comparable across models. The full schema is given in Appendix~\ref{app:action-schema}.

\subsection{Outcome rewards and offline objectives}
\label{sec:method-objectives}

We compare four offline policy-learning objectives using the same backbone,
prompts, action schema, decoding constraints, and fine-tuning setup. The
backbone is Qwen-2.5-7B-Instruct, fine-tuned with QLoRA. Each input is rendered
as a historical state for a program-window pair, and the target is the JSON
serialization of the trials launched in that window. The first three objectives differ only
in how examples are weighted during training. The fourth, implicit Q-learning, instead learns a value function from the logged outcomes and weights each example by the estimated advantage of its action, so it can favor decisions that beat the typical outcome from a comparable state rather than simply up-weighting high-reward windows.

The first objective is behavioral cloning. Every example receives weight $1.0$, and the model is trained to imitate the historical sponsor action distribution. This provides an unweighted supervised fine-tuning baseline that does not use downstream outcome information.

The second objective is reward-weighted behavioral cloning, an offline RL-style objective inspired by advantage-weighted regression. Each launched trial $i$ receives the per-trial reward
\begin{equation}
r_i = 2\,a_i - f_i + \log(1+\mathrm{rev}_i),
\label{eq:reward}
\end{equation}
where $a_i\!=\!1$ if trial $i$ became pivotal in an FDA approval and $f_i\!=\!1$ if it failed or was terminated. $\mathrm{rev}_i$ is the drug's peak annual revenue attributed to trial $i$: the total is split across the program's FDA approvals, and each approval's share is then divided among the pivotal trials that supported it. The window score $R_{p,t}$ is the mean of $r_i$ over the trials launched in that window, a direct attribution of realized regulatory and commercial outcomes to the trials that produced them. We standardize $R_{p,t}$ within each split and convert it into a clipped sampling weight $w_{p,t}=\mathrm{clip}(0.5+\tilde{R}_{p,t},0.1,3.0)$, training on the rebalanced dataset so higher-reward windows are sampled more often. We analyze the reward terms and this per-trial credit-assignment choice in Appendix~\ref{app:reward-ablation}.

The third objective is learned-reward-weighted behavioral cloning. Rather than the fixed coefficients of Eq.~\ref{eq:reward}, we \emph{learn} the reward (Appendix~\ref{app:learned_reward} ) by regressing trial-design features onto the same per-trial outcome target, so the model infers the objective from the structure of the trials instead of from a rule-based formula. The fitted per-trial scores are then turned into sampling weights by the same method as the rule-based reward.

The fourth objective is value-based offline reinforcement learning with implicit Q-learning (IQL). While the previous objectives reweight imitation by an \emph{observed} reward, IQL weights examples by an \emph{estimated} advantage, giving policy improvement beyond behavioral cloning. We treat each launched trial as an action and fit an IQL value model over all programs: a state-value function $V_\psi(s)$ trained by expectile regression with expectile $\tau=0.7$ (so $V_\psi$ tracks an upper quantile of return rather than its mean) and an action-value $Q_\phi(s,a)=R_{p,t}+\gamma\,V_\psi(s')$ with discount $\gamma=0.95$ that bootstraps the discounted per-trial reward \citep{kostrikov2021offline}. Each training example is then weighted by an advantage-weighted transform of $A(s,a)=Q_\phi(s,a)-V_\psi(s)$, which measures how much a launch decision beat the field-average value from a comparable program state, at inverse temperature $\beta=0.5$, with per-trial advantages aggregated to a window weight under the same clip-and-oversample scheme used for the previous rewards.

\subsection{Baselines and evaluation}
\label{sec:method-evaluation}

Our primary comparison is the four fine-tuned models against four tool-using agents (GPT-5.4, Claude Opus 4.5, Gemini 3.1 Pro, and Qwen3-235B-Thinking-2507) on a shared scaffold. We also included prompt-only approaches for Qwen-2.5-7B-Instruct (backbone for the trained models), GPT-5.4 and a random baseline. All approaches return the same 14-field action schema. The fine-tuned models and the prompt-only LLMs are run with a structured summary of the program, its history, and wider context at the decision date (Appendix~\ref{app:full-state-prompt}). Each agent receives the drug identifier and decision date plus 10 typed retrieval tools covering sponsor trial history, competitor and indication landscape, SEER epidemiology, sponsor revenue, and document retrieval of FDA reviews, EMA EPARs, and publications from OpenAlex and PubMed. Each tool is date-gated, so information dated on or after the decision window is not returned. Prompt-format ablations for the prompt-only baselines are reported in Appendix~\ref{app:no-tools-agent}.

Each predicted portfolio is scored against the held-out ground-truth portfolio by greedy one-to-one pairing. Free-text fields (indication, trial description, comparator, and combination partners) are matched by embedding similarity, and categorical or enumerated fields by exact match; this rule applies to all metrics below, and credits same-disease paraphrases that lexical string matching would miss (Appendix~\ref{app:metric_implementation}).

We report the following metrics. \textit{Indication F1} is the standard F1 over indication matches: precision is matched pairs divided by predicted trials, recall is matched pairs divided by ground-truth trials. It measures whether the model identifies the correct disease and biomarker setting. \textit{Strict F1} measures wider-trial alignment; a pair is formed only if indication, phase, and strategy all match. \textit{Field-match \%} is, for an indication paired trial, the percentage of the remaining 13 fields that match, averaged over matched pairs. 

Beyond next-window matching, we analyzed whether models reproduce the designs of trials that historically succeeded more than those that failed. Across the five single-drug holdouts, we take every substantive held-out trial with a known outcome (53 successes and 76 failures). For each, we test whether the model proposes a matching trial within one window of its decision point, where a match requires the full design to agree on indication, phase, and strategy. We report the match rate on successful and on failed trials and their difference, with a 2000-sample bootstrap 95\% CI. A positive difference means the model reproduces winning designs more than losing ones.

To probe robustness to compounding state error, we additionally evaluate the offline-trained models, Qwen-2.5-7B, GPT-5.4, and the GPT-5.4 agent in an autoregressive rollout on the five single-drug holdouts. In this setting, each model sees its own earlier predicted launches as part of the launch history for later windows, replacing the oracle history given by the static evaluation. We report the change in metrics from the static to the autoregressive evaluation as a measure of how much each method's accuracy degrades when conditioning on its own outputs rather than ground truth.

Finally, to assess whether agent retrieval and offline policy learning compose, we ran a hybrid evaluation in which the GPT-5.4 agent's tool calls and the results from its date-gated retrieval for each window are serialized as a retrieved-evidence block, appended to the same prompt the models were trained on. The fine-tuned models then produce the same 14-field action schema.

\begin{figure}[!t]
  \centering
  \includegraphics[width=\linewidth]{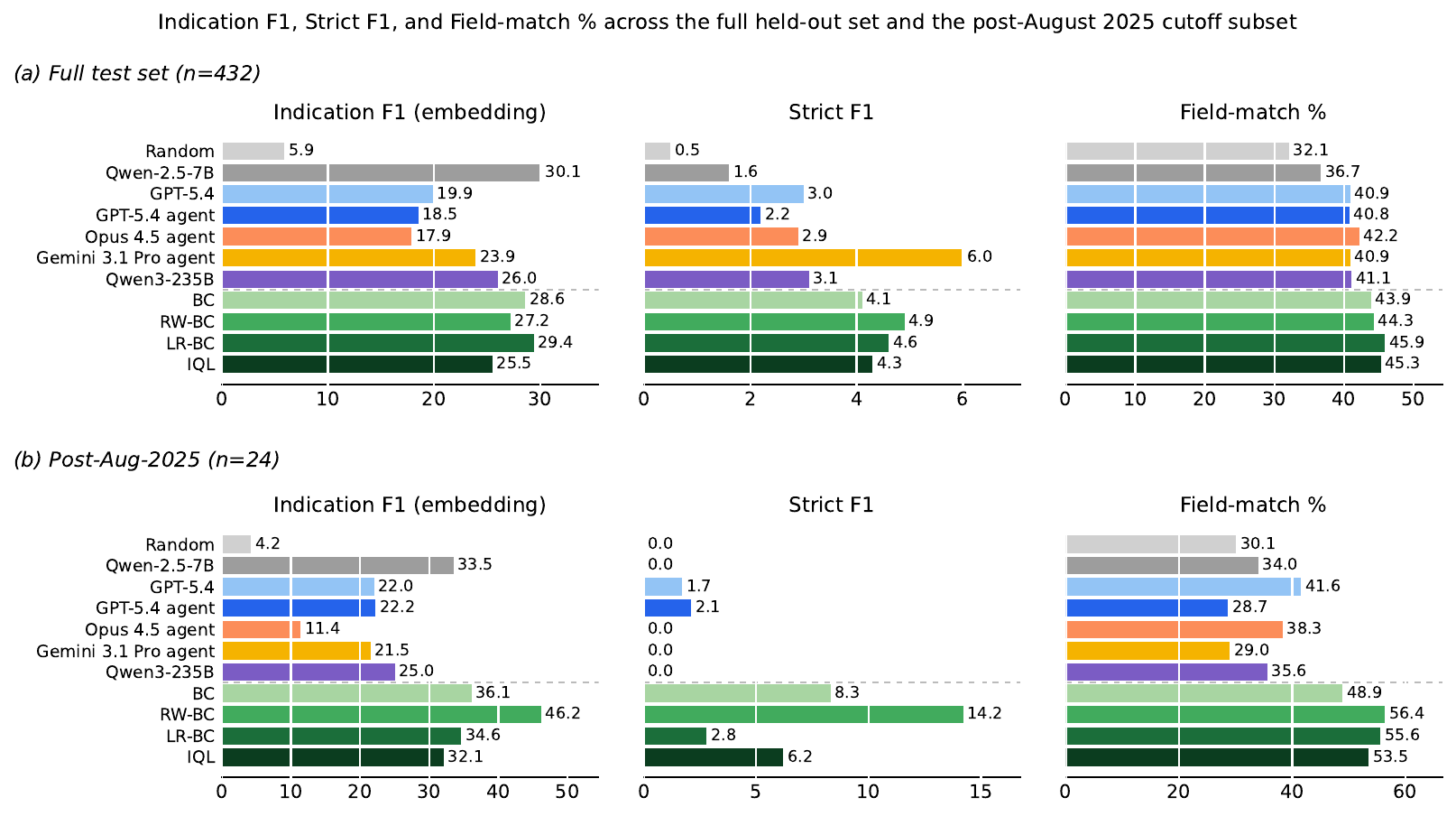}
  \caption{Results are reported as windows-weighted means across the eight held-out evaluation splits. [a] is the union of all $n{=}432$ held-out windows, while [b] is the cross-split subset of $n{=}24$ windows whose decision date falls after August 2025, the most recent knowledge cutoff for the models evaluated here. BC, RW-BC, LR-BC, and IQL denote behavioral cloning, reward-weighted behavioral cloning, learned-reward behavioral cloning, and value-based implicit Q-learning.}
  \label{fig:headline}
\end{figure}

\begin{table}[!t]
  \caption{Indication F1 by holdout split}
  \label{tab:per-split}
  \vskip 0.05in
  \centering
  \footnotesize
  \setlength{\tabcolsep}{4pt}
  \resizebox{\columnwidth}{!}{%
  \begin{tabular}{@{}lr|rrrrr|rrrr@{}}
    \toprule
                                &     & \multicolumn{5}{c}{Baseline + tool agents}                                          & \multicolumn{4}{c}{Offline-trained models}                       \\
    \cmidrule(lr){3-7}\cmidrule(lr){8-11}
    Split                       & $n$ & Qwen-2.5-7B            & GPT-5.4         & Gemini 3.1 Pro  & Opus 4.5        & Qwen3-235B      & BC              & RW-BC           & LR-BC           & IQL             \\
    \midrule
    tagrisso                    & 25  & 59.3            & 55.1            & 44.9            & 48.1            & \textbf{61.2}   & 54.5            & 53.0            & 53.6            & 41.1            \\
    gleevec                     & 35  & 29.8            & 10.0            & \phantom{0}7.7  & \phantom{0}9.2  & \phantom{0}7.0  & \textbf{39.9}   & 33.9            & 33.4            & 38.7            \\
    imbruvica                   & 19  & 21.2            & \phantom{0}5.1  & \phantom{0}2.5  & \phantom{0}1.6  & 21.4            & 34.1            & \textbf{39.2}   & 36.9            & 25.6            \\
    keytruda                    & 26  & 30.5            & 13.0            & 16.9            & 10.2            & 26.2            & 19.1            & 25.0            & \textbf{32.3}   & 16.0            \\
    opdivo                      & 31  & 27.2            & 17.7            & 23.4            & 21.1            & 23.3            & 19.6            & 25.2            & 20.8            & \textbf{32.1}   \\
    \midrule
    AstraZeneca sponsor         & 96  & 32.3            & 22.8            & 31.1            & 22.8            & \textbf{32.8}   & 27.9            & 22.4            & 26.7            & 24.4            \\
    Drug class (PD-1/L1)        & 155 & \textbf{25.9}   & 13.4            & 23.0            & 15.8            & 22.9            & 23.0            & 18.4            & 23.1            & 17.7            \\
    \midrule
    post-Sep 2024 (Qwen-2.5 cutoff)         & 44  & 29.6            & 21.9            & 22.3            & 13.5            & 20.9            & 35.8            & \textbf{45.6}   & 41.3            & 36.9            \\
    \dag\, post-Aug 2025 (all-model cutoff) & 24  & 33.5            & 22.2            & 21.5            & 11.4            & 25.0            & 36.1            & \textbf{46.2}   & 34.6            & 32.1            \\
    \bottomrule
  \end{tabular}%
  }
\end{table}

\section{Results and Discussion}
\label{sec:results}

Figure~\ref{fig:headline} shows results on both the full held-out set and the contamination-free post-August 2025 subset, on which no model could have memorized the decisions. The offline-trained models outperform the non-fine-tuned baselines, particularly on the post-August 2025 subset, where the strongest model, reward-weighted behavioral cloning (RW-BC), reaches 46.2\% indication F1 and 14.2\% strict F1 versus 25.0\% and 2.1\%, respectively, for the best-performing tool agent on each metric. The gap is robust to the small sample. In a paired bootstrap over shared windows RW-BC significantly exceeds the strongest agent on Indication F1 ($\Delta\,{+}21.1$, 95\% CI $[1.7, 41.1]$, $p{=}0.017$), every fine-tuned model beats all four frontier agents on indication and field-match ($P\!\approx\!0.99\text{--}1.0$), and they beat the base model on field-match. The agentic setup did not show differentiated performance vs the one-shot counterpart for GPT-5.4. The reduction in the agents' scores on the post-cutoff subset suggests they rely on pre-training knowledge where it is available, whereas the fine-tuned models learn a decision mapping from the offline episodes.

The offline trained models show particularly strong performance in the Strict F1 metric, which rewards broader trial design decisions over just indication prediction (Figure~\ref{fig:headline}, Appendix Table~\ref{tab:headline-embed}). The results of the base model illustrates why Indication F1 alone can be misleading. On the Tagrisso holdout, it predicted NSCLC for 100\% of the trials, falling back to its parametric knowledge of the drug's flagship indication and obtaining a good score of 59.3. A similar strategy on the post-cutoff subset gave it a competitive 33.5 Indication F1 by over-predicting common diseases, but it scored 0.0 in Strict F1, showing no ability to combine an indication with other trial design decisions. The fine-tuned models instead aligned their indication matches with the wider decisions made. The low absolute Strict F1 values are expected, since the historical action is one realized decision among multiple plausible counterfactuals. We therefore treat the metrics as alignment with historical sponsor behavior rather than proof of optimality.

Per-split breakdowns (Table~\ref{tab:per-split}) show that the gap between trained models and baselines is widest on the less-iconic programs such as gleevec and imbruvica, where the agents collapse to single digits while the offline models maintain performance in the 25--40\% range. On heavily-documented programs such as tagrisso, all models performed well. No single fine-tuned model dominates every split, but RW-BC leads on aggregate. This is consistent with prior work showing that behavioral cloning can match or beat offline value-based RL when data are limited \citep{kumar2022when,emmons2022rvs,brandfonbrener2022when,wang2020critic}.

We next asked whether models reproduced the designs of trials that succeeded more than those that failed (Table~\ref{tab:succ-fail}). The base model and the tool agents reproduce failed designs more than successful ones (Qwen-2.5-7B $-39.3$ and GPT-5.4 $-17.5$, both with 95\% CIs excluding zero). Only the reward-trained models reverse this trend (LR-BC $+11.0$, RW-BC $+3.7$; positive, though not individually significant), demonstrating the value of weighting individual trials over base behavioral cloning.

\begin{table}[!t]
  \caption{Success versus failed trial design matches on the five single-drug holdouts.}
  \label{tab:succ-fail}
  \vskip 0.05in
  \centering
  \small
  \setlength{\tabcolsep}{3.5pt}
  \resizebox{\columnwidth}{!}{%
  \begin{tabular}{@{}lrrc@{}}
    \toprule
    Model            & success\% & failed\% & discrim [95\% CI]              \\
    \midrule
    LR-BC            & 18.9     & \phantom{0}7.9  & $+11.0$ $[-0.5, 23.0]$  \\
    RW-BC            & 20.8     & 17.1            & $+3.7$ $[-9.9, 18.4]$      \\
    IQL              & 18.9     & 23.7            & $-4.8$ $[-19.3, 9.1]$      \\
    Gemini 3.1 Pro agent & \phantom{0}3.8 & 11.8 & $-8.0$ $[-16.5, 0.4]$      \\
    BC               & 15.1     & 25.0            & $-9.9$ $[-23.1, 4.4]$      \\
    Opus 4.5 agent   & \phantom{0}5.7 & 15.8       & $-10.1$ $[-19.9, -0.3]$    \\
    GPT-5.4 agent    & \phantom{0}7.5 & 25.0       & $-17.5$ $[-29.1, -5.4]$   \\
    Qwen3-235B agent & 11.3     & 35.5            & $-24.2$ $[-37.9, -10.5]$   \\
    Qwen-2.5-7B      & \phantom{0}9.4 & 48.7            & $-39.3$ $[-53.0, -25.0]$   \\
    \bottomrule
  \end{tabular}}
\end{table}

Static evaluation gives each model the true historical launch history at every window. Autoregressive evaluation instead feeds each model its own previous predictions, testing sensitivity to compounding errors (Table~\ref{tab:autoregressive}). Most of the fine-tuned models were robust to this setup, with BC and RW-BC returning similar results. LR-BC and the base model saw a large drop in Indication F1 ($-5.8$ and $-9.3$ respectively), while GPT-5.4 returned the lowest absolute score ($17.4$). Across metrics, the fine-tuned models remained above the baselines, showing that the strategic direction of drug development can be trained into the model's weights, enabling them to be robust to their own errors.

\begin{table}[!t]
  \caption{Static versus autoregressive (AR) performance on the five single-drug holdouts.}
  \label{tab:autoregressive}
  \vskip 0.1in
  \centering
  \footnotesize
  \setlength{\tabcolsep}{3pt}
  \resizebox{\columnwidth}{!}{%
  \begin{tabular}{@{}l rrr rrr rrr@{}}
    \toprule
                     & \multicolumn{3}{c}{Indication F1} & \multicolumn{3}{c}{Strict F1} & \multicolumn{3}{c}{Field-match \%} \\
    \cmidrule(lr){2-4}\cmidrule(lr){5-7}\cmidrule(lr){8-10}
    Method           & Static & AR    & $\Delta$ & Static & AR    & $\Delta$ & Static & AR            & $\Delta$        \\
    \midrule
    BC               & 33.4   & 34.6  & $+1.2$   & 3.6    & 4.1   & $+0.4$   & 47.6   & 48.1          & $+0.4$          \\
    RW-BC            & 35.3   & 36.8  & $+1.5$   & 5.2    & 4.0   & $-1.2$   & 46.7   & 46.0          & $-0.7$          \\
    LR-BC            & 35.4   & 29.6  & $-5.8$   & 8.2    & 3.5   & $-4.7$   & 48.0   & 44.4          & $-3.5$          \\
    IQL              & 30.7   & 27.6  & $-3.1$   & 3.6    & 3.8   & $+0.2$   & 46.2   & 46.6          & $+0.4$          \\
    \midrule
    Qwen-2.5-7B      & 33.6   & 24.3  & $-9.3$   & 1.9    & 1.2   & $-0.7$   & 38.7   & 37.3          & $-1.4$          \\
    GPT-5.4 agent    & 20.0   & 17.4  & $-2.6$   & 2.1    & 2.1   & \phantom{$+$}$0.0$   & 41.3   & 37.3          & $-4.0$          \\
    \bottomrule
  \end{tabular}%
  }
\end{table}

Combining an offline policy with agent-retrieved evidence as a naive offline-to-online composition improved the strongest fine-tuned model (RW-BC) to 59.0\% Indication F1 and 18.6\% Strict F1 on the post-cutoff subset. Given the small sample size this gain is only suggestive but points to retrieval-augmented offline policies as a promising future direction.

This work has several limitations. First, the rewards are retrospective associations, not causal estimates of the value of a trial-launch decision. Approval and revenue depend on drug quality, sponsor resources, pricing, and competition that the public state only partly captures. Because decisions are also affected by private information, there are unobserved confounders, and no objective we tried, including value-based IQL, cleanly separates a decision's own quality from the program trajectory (Appendix~\ref{app:reward-ablation}). Per-trial credit is also local. A trial is penalized only for its own failure, so an early trial that was later abandoned in a subsequent phase is scored roughly neutrally; propagating a program's eventual failure back to its enabling trials is future work, since naive back-propagation re-blames trials indiscriminately. Second, the post-August 2025 contamination-controlled subset is small ($n\!=\!24$), so it is a useful comparison but not a definitive estimate of prospective performance. Finally, the schema ensures comparable outputs but abstracts away details that matter in real trial design, including endpoint definitions and operational feasibility. We therefore read this as evidence that historical development data can contain a learnable strategic signal enabling agents to plan biomedical experiments, not a claim that the learned policy is optimal.

These results motivate more work on closed-loop agentic systems and offline-to-online methods. A north star would be a clinical-development simulator that supports counterfactual rollouts to turn the retrospective reward into a causal one and enable online adaptation. Nearer-term steps include training the offline objective on retrieval-augmented prompts using already-collected agent transcripts, so that hypothesis generation, evidence retrieval, and structured action planning are learned jointly. Furthermore, new sponsor decisions can be incorporated as fresh episodes, and hierarchical RL with sub-goals could be used to tackle the problems of long horizons and sparse delayed rewards \citep{shin2023guide,levine2020offline,prudencio2023survey}. Formulating, planning, and acting on clinical-development strategy remains a difficult sequential decision-making problem that frontier agents do not yet solve.

\newpage
\section*{Code and data availability}

Code available at \url{https://github.com/WilliamBolton/clinical-trial-strategy-agents}. The structured dataset is available at \url{https://huggingface.co/datasets/WillBolton/oncology-trial-strategy}.

\section*{Impact Statement}
\label{sec:impact}

This paper presents a research artifact for studying clinical-development strategy from public oncology data. The models are intended for retrospective structured forecasting and evaluation, not for autonomous trial design, regulatory submission, investment decisions, or patient care. The main risks are misuse and bias amplification, since public development histories overrepresent large sponsors, marketed assets, common indications, and strategies that leave visible regulatory or commercial traces. Models trained on this history may therefore reproduce incumbent development patterns and underrepresent rare, pediatric, low-resource, or commercially weaker indications. We mitigate these risks by using only public sources, date-gating evidence at the decision window, evaluating across drug, sponsor, drug-class, and temporal splits, and treating outputs as program-level research predictions rather than clinical or commercial recommendations. No patient-level records, confidential sponsor materials, or proprietary information was used in this research.

\section*{Acknowledgements}
This work was funded by ARIA, DSIT and Pillar VC under the Encode: AI for Science Fellowship.

\bibliography{paper}

\begin{thebibliography}{17}
\providecommand{\natexlab}[1]{#1}
\providecommand{\url}[1]{\texttt{#1}}
\expandafter\ifx\csname urlstyle\endcsname\relax
  \providecommand{\doi}[1]{doi: #1}\else
  \providecommand{\doi}{doi: \begingroup \urlstyle{rm}\Url}\fi

\bibitem[Brandfonbrener et~al.(2022)Brandfonbrener, Bietti, Buckman, Laroche,
  and Bruna]{brandfonbrener2022when}
Brandfonbrener, D., Bietti, A., Buckman, J., Laroche, R., and Bruna, J.
\newblock When does return-conditioned supervised learning work for offline
  reinforcement learning?
\newblock In \emph{Advances in Neural Information Processing Systems
  (NeurIPS)}, 2022.

\bibitem[Chen et~al.(2025)Chen, Hu, Cai, Lu, Wang, Cao, Lin, Xu, Wu, Cao,
  et~al.]{chen2025trialbench}
Chen, J., Hu, Y., Cai, M., Lu, Y., Wang, Y., Cao, X., Lin, M., Xu, H., Wu, J.,
  Cao, X., et~al.
\newblock Trialbench: Multi-modal ai-ready datasets for clinical trial
  prediction.
\newblock \emph{Scientific Data}, 12\penalty0 (1):\penalty0 1564, 2025.

\bibitem[Chen et~al.(2021)Chen, Lu, Rajeswaran, Lee, Grover, Laskin, Abbeel,
  Srinivas, and Mordatch]{chen2021decision}
Chen, L., Lu, K., Rajeswaran, A., Lee, K., Grover, A., Laskin, M., Abbeel, P.,
  Srinivas, A., and Mordatch, I.
\newblock Decision transformer: Reinforcement learning via sequence modeling.
\newblock \emph{Advances in neural information processing systems},
  34:\penalty0 15084--15097, 2021.

\bibitem[Emmons et~al.(2022)Emmons, Eysenbach, Kostrikov, and
  Levine]{emmons2022rvs}
Emmons, S., Eysenbach, B., Kostrikov, I., and Levine, S.
\newblock {RvS}: What is essential for offline {RL} via supervised learning?
\newblock In \emph{International Conference on Learning Representations
  (ICLR)}, 2022.

\bibitem[Fu et~al.(2022)Fu, Huang, Xiao, Glass, and Sun]{fu2022hint}
Fu, T., Huang, K., Xiao, C., Glass, L.~M., and Sun, J.
\newblock Hint: Hierarchical interaction network for clinical-trial-outcome
  predictions.
\newblock \emph{Patterns}, 3\penalty0 (4), 2022.

\bibitem[Huang et~al.(2025)Huang, Zhang, Wang, Qu, Lu, Roohani, Li, Qiu, Li,
  Zhang, et~al.]{huang2025biomni}
Huang, K., Zhang, S., Wang, H., Qu, Y., Lu, Y., Roohani, Y., Li, R., Qiu, L.,
  Li, G., Zhang, J., et~al.
\newblock Biomni: A general-purpose biomedical ai agent.
\newblock \emph{biorxiv}, 2025.

\bibitem[Janner et~al.(2021)Janner, Li, and Levine]{janner2021offline}
Janner, M., Li, Q., and Levine, S.
\newblock Offline reinforcement learning as one big sequence modeling problem.
\newblock \emph{Advances in neural information processing systems},
  34:\penalty0 1273--1286, 2021.

\bibitem[Jin et~al.(2024)Jin, Wang, Floudas, Chen, Gong, Bracken-Clarke, Xue,
  Yang, Sun, and Lu]{jin2024matching}
Jin, Q., Wang, Z., Floudas, C.~S., Chen, F., Gong, C., Bracken-Clarke, D., Xue,
  E., Yang, Y., Sun, J., and Lu, Z.
\newblock Matching patients to clinical trials with large language models.
\newblock \emph{Nature communications}, 15\penalty0 (1):\penalty0 9074, 2024.

\bibitem[Kostrikov et~al.(2021)Kostrikov, Nair, and
  Levine]{kostrikov2021offline}
Kostrikov, I., Nair, A., and Levine, S.
\newblock Offline reinforcement learning with implicit q-learning.
\newblock \emph{arXiv preprint arXiv:2110.06169}, 2021.

\bibitem[Kumar et~al.(2022)Kumar, Hong, Singh, and Levine]{kumar2022when}
Kumar, A., Hong, J., Singh, A., and Levine, S.
\newblock When should we prefer offline reinforcement learning over behavioral
  cloning?
\newblock In \emph{International Conference on Learning Representations
  (ICLR)}, 2022.
\newblock arXiv:2204.05618.

\bibitem[Levine et~al.(2020)Levine, Kumar, Tucker, and Fu]{levine2020offline}
Levine, S., Kumar, A., Tucker, G., and Fu, J.
\newblock Offline reinforcement learning: Tutorial, review, and perspectives on
  open problems.
\newblock \emph{arXiv preprint arXiv:2005.01643}, 2020.

\bibitem[Nair et~al.(2020)Nair, Gupta, Dalal, and Levine]{nair2020awac}
Nair, A., Gupta, A., Dalal, M., and Levine, S.
\newblock Awac: Accelerating online reinforcement learning with offline
  datasets.
\newblock \emph{arXiv preprint arXiv:2006.09359}, 2020.

\bibitem[Peng et~al.(2019)Peng, Kumar, Zhang, and Levine]{peng2019awr}
Peng, X.~B., Kumar, A., Zhang, G., and Levine, S.
\newblock Advantage-weighted regression: Simple and scalable off-policy
  reinforcement learning.
\newblock \emph{arXiv preprint arXiv:1910.00177}, 2019.

\bibitem[Prudencio et~al.(2023)Prudencio, Maximo, and
  Colombini]{prudencio2023survey}
Prudencio, R.~F., Maximo, M.~R., and Colombini, E.~L.
\newblock A survey on offline reinforcement learning: Taxonomy, review, and
  open problems.
\newblock \emph{IEEE transactions on neural networks and learning systems},
  35\penalty0 (8):\penalty0 10237--10257, 2023.

\bibitem[Schmidgall et~al.(2024)Schmidgall, Ziaei, Harris, Reis, Jopling, and
  Moor]{schmidgall2024agentclinic}
Schmidgall, S., Ziaei, R., Harris, C., Reis, E., Jopling, J., and Moor, M.
\newblock Agentclinic: a multimodal agent benchmark to evaluate ai in simulated
  clinical environments.
\newblock \emph{arXiv preprint arXiv:2405.07960}, 2024.

\bibitem[Shin \& Kim(2023)Shin and Kim]{shin2023guide}
Shin, W. and Kim, Y.
\newblock Guide to control: Offline hierarchical reinforcement learning using
  subgoal generation for long-horizon and sparse-reward tasks.
\newblock In \emph{IJCAI}, pp.\  4217--4225, 2023.

\bibitem[Wang et~al.(2020)Wang, Novikov, Zolna, Springenberg, Reed, Shahriari,
  Siegel, Merel, Gulcehre, Heess, and de~Freitas]{wang2020critic}
Wang, Z., Novikov, A., Zolna, K., Springenberg, J.~T., Reed, S., Shahriari, B.,
  Siegel, N., Merel, J., Gulcehre, C., Heess, N., and de~Freitas, N.
\newblock Critic regularized regression.
\newblock In \emph{Advances in Neural Information Processing Systems
  (NeurIPS)}, 2020.

\end{thebibliography}
\bibliographystyle{icml2026}

\appendix
\onecolumn
\raggedbottom

\section{Supplementary Methods}
\label{app:supp_methods}

\subsection{Action schema}
\label{app:action-schema}

Every predicted and ground-truth trial in our benchmark conforms to a JSON Schema with 14 required fields, enumerated vocabularies, and a portfolio-size cap of 10 trials per decision window. The schema is enforced at decode time by vLLM's \texttt{GuidedDecodingParams} for the locally served Qwen-2.5-7B full and the offline-trained models. For the prompted-LLM baselines, including the Qwen API and the GPT-5.4 \textit{min}, \textit{full}, and \textit{agent} variants, the schema is described in the system prompt and JSON output is parsed post-hoc with brace-balanced extraction and truncation repair. The 14 fields and their enumerated vocabularies are listed in Table~\ref{tab:app-schema-fields}, and an example portfolio entry is shown in Listing~\ref{lst:app-schema-example}.

\begin{table}[H]
  \caption{Action schema. Every predicted trial is a 14-field structured object. Categorical fields are constrained to a fixed enum at decode time. List fields (combination partners, secondary endpoint types) are open-vocabulary. The comparator field accepts a string or \texttt{null}.}
  \label{tab:app-schema-fields}
  \centering
  \small
  \setlength{\tabcolsep}{4pt}
  \begin{tabular}{@{}llp{0.45\columnwidth}@{}}
    \toprule
    Field                       & Type                & Vocabulary or notes                                                          \\
    \midrule
    \texttt{indication}         & string              & free-text disease label                                                      \\
    \texttt{trial\_description} & string              & free-text trial summary                                                      \\
    \texttt{phase}              & enum                & Phase~1, Phase~2, Phase~3, Phase~4                                          \\
    \texttt{strategy}           & enum                & confirmatory, indication\_expansion, combination, dose, geographic, rwe      \\
    \texttt{study\_type}        & enum                & interventional, observational                                                \\
    \texttt{allocation}         & enum                & randomized, non\_randomized, na                                              \\
    \texttt{masking}            & enum                & open\_label, blinded                                                         \\
    \texttt{n\_arms\_bucket}    & enum                & 1, 2, 3+                                                                     \\
    \texttt{enrollment\_bucket} & enum                & small, mid, large                                                            \\
    \texttt{combo\_partners}    & list of strings     & open-vocabulary (drug names)                                                 \\
    \texttt{comparator}         & string \textbar{} null & free-text or null                                                         \\
    \texttt{region}             & enum                & single\_country, multi\_regional                                             \\
    \texttt{primary\_endpoint\_type} & enum            & ORR, PFS, OS, DLT, safety, other                                             \\
    \texttt{secondary\_endpoint\_types} & list of strings & open-vocabulary                                                          \\
    \bottomrule
  \end{tabular}
\end{table}

\begin{figure}[H]
\begin{verbatim}
{
  "indication": "Non-Small Cell Lung Cancer",
  "trial_description":
      "Pembrolizumab vs platinum chemotherapy in 1L NSCLC",
  "phase": "Phase 3",
  "strategy": "indication_expansion",
  "study_type": "interventional",
  "allocation": "randomized",
  "masking": "open_label",
  "n_arms_bucket": "2",
  "enrollment_bucket": "large",
  "combo_partners": [],
  "comparator": "platinum doublet chemotherapy",
  "region": "multi_regional",
  "primary_endpoint_type": "PFS",
  "secondary_endpoint_types": ["OS", "ORR"]
}
\end{verbatim}
\caption{An example trial conforming to the action schema. Real predictions and ground-truth trials follow this shape exactly. The surrounding portfolio is a list of one to ten such objects.}
\label{lst:app-schema-example}
\end{figure}

\subsection{Primary data sources and auxiliary corpus}
\label{app:data_sources}

The benchmark state, action, and reward are constructed from five primary
public-source streams. ClinicalTrials.gov provides trials, eligibility criteria, outcomes, and trial timing. Drugs@FDA provides review PDFs used by the approval linker. SEC EDGAR provides 10-K, 10-Q, 6-K, and 20-F filings parsed for quarterly product revenue. CMS Part D provides a patient-reach signal. SEER provides US cancer epidemiology.

The same ingestion pipeline also collects auxiliary sources, including EMA EPARs, DailyMed, OpenAlex, PubMed, GDELT, AACT, ESMO, PatentsView, and FDA designations. These sources populate the document corpus available to the LLM tool-using baseline, but they do not contribute to the behavioral-cloning, reward-weighted, or learned-reward training state or reward.

Raw fetches are stored immutably with content-hash addressing and full provenance. Each fetch records the source URL, fetch timestamp, content type, and source identifier. Fetches are then normalized into trial records, document records, per-trial competitive snapshots evaluated as of each trial start date, and quarterly revenue entries.

\subsection{Parsing and temporal snapshot construction}
\label{app:parsing_snapshots}

ClinicalTrials.gov entries are converted into trial records containing arms, eligibility criteria, outcome measures, dates, references, and registry status. Drugs@FDA review PDFs are parsed into document records with extracted text and metadata. SEC filings are parsed from HTML and used to extract quarterly drug-level revenue where available. CMS Part D records are used to derive beneficiary counts. SEER tables provide indication-level epidemiological
context.

For each sponsor trial start date, the pipeline builds a competitive snapshot using only information available up to that date. The snapshot records indication-relevant competitor trials, prior approvals, treatment alternatives, and epidemiology. Later events are excluded from the snapshot used to construct the state.

\subsection{Approval linkage}
\label{app:approval_linkage}

FDA approval rewards require linking approvals to the trials that supported them. The linker first searches FDA review text for direct NCT identifier mentions. It then applies a trial-acronym map built from ClinicalTrials.gov brief titles and acronym fields. This handles variants such as \texttt{KEYNOTE-024}, \texttt{KEYNOTE 024}, and \texttt{keynote024}. If direct identifier and acronym matching fail, the linker falls back to sponsor trials in the approved indication, prioritizing phase 3 or later trials within a window before approval.

\subsection{Reward normalization and sampling weights}
\label{app:reward_weights}

Observed rewards are normalized within each training split before being converted into sampling weights. For a program-window pair $(p,t)$, the normalized reward $\tilde{R}_{p,t}$ is converted into a clipped sampling weight:
\begin{equation}
w_{p,t} =
\mathrm{clip}\!\left(0.5+\tilde{R}_{p,t},\,0.1,\,3.0\right).
\label{eq:app_reward_weight}
\end{equation}

The lower bound keeps low-reward windows present with non-zero probability. The upper bound prevents a small number of high-reward windows from dominating the fine-tuning dataset.

Reward-weighted training is implemented through stochastic oversampling. Each example is included $\lfloor w_{p,t}\rfloor$ times, with one additional copy sampled with probability $w_{p,t}-\lfloor w_{p,t}\rfloor$. The resulting dataset is then trained with the same next-token prediction objective as behavioral cloning.

\subsection{Fine-tuning implementation}
\label{app:training_details}

All fine-tuned models use Qwen-2.5-7B-Instruct with QLoRA. States are rendered as the prompt by \texttt{build\_temporal\_prompt}. QLoRA uses 4-bit NF4 quantization, LoRA rank $r=64$, $\alpha=128$, dropout 0.05, and target modules \texttt{q\_proj}, \texttt{k\_proj}, \texttt{v\_proj}, \texttt{o\_proj}, \texttt{gate\_proj}, \texttt{up\_proj}, and \texttt{down\_proj}.

Training uses 8 epochs, per-device batch size 4, gradient accumulation 4, effective batch size 16, learning rate $3{\times}10^{-4}$, cosine scheduling, 3 percent warmup, bf16, maximum sequence length 4096, a 10 percent held-out validation split, and random seed 42. Optimization uses \texttt{adamw\_8bit} through Unsloth or \texttt{adamw\_torch} through Hugging Face, with TRL \texttt{SFTTrainer}.

\subsection{Learned reward model}
\label{app:learned_reward}

For the learned-reward objective we replace the rule-based reward of Eq.~\ref{eq:reward} with one learned from data. Each trial is mapped to a feature vector summarizing its design (phase, strategy, study type, enrollment, combination, endpoint, and novelty) together with program context (maturity, indication breadth, competitive pressure, and prior outcomes). A regression model is fit to predict the per-trial reward target of Eq.~\ref{eq:reward} from these features, so the learned reward captures which design choices tend to precede good outcomes rather than encoding them by hand. Its per-trial predictions are averaged over each window, standardized, and clipped into sampling weights exactly as for the rule-based reward (Eq.~\ref{eq:app_reward_weight}); training is otherwise identical.

\subsection{Prompt}
\label{app:full-state-prompt}

The prompt contains, for the program at the decision date:
\begin{itemize}
\item drug name, mechanism of action, and target;
\item completed and ongoing sponsor trials, each with phase, start and completion dates, enrolment, and outcome label where available;
\item prior FDA approvals and their indications;
\item the current competitive landscape: other sponsors' active trials in the same indication, MoA-enriched, with CMS Part D spending where available;
\item indication alternatives (treatments competing for the same patient population);
\item SEER incidence and 5-year survival for each indication of interest;
\item the last eight quarters of sponsor product revenue;
\item current market position (revenue rank, share trajectory).
\end{itemize}
None of these fields contain information dated on or after the decision window. The same prompt is shown to the fine-tuned models at training and inference time, and to the prompt-only LLM baselines that are run in ``full'' mode.

\subsection{Inference and baseline implementation}
\label{app:inference_details}
\label{app:baseline_implementation}

Local inference uses vLLM 0.7.3 with bf16, \texttt{gpu\_memory\_utilization=0.85}, and \texttt{max\_model\_len=16384}. LoRA serving uses \texttt{enable\_lora=True} and \texttt{max\_lora\_rank=64}. Local models use temperature-zero decoding and the JSON-schema-guided decoder.

GPT-5.4 baselines run through the OpenAI API at temperature zero. Tool-using agents on backbones other than GPT-5.4 (Claude Opus 4.5, Gemini 3.1 Pro, Qwen3-235B-Thinking-2507) are routed through OpenRouter so the same agent scaffold and 10-tool retrieval interface can be applied uniformly across backbones. For all API-served models the schema is described in the system prompt with a worked example rather than enforced by the local constrained-decoding stack, and a lightweight parser removes Markdown fences, extracts the largest balanced JSON object, and attempts simple truncation repair. API cache keys hash the state, candidate set, evaluation mode, and prompt version, so re-running an evaluation incurs no API cost.

The random baselines use empirical distributions from the training data. The mode-random baseline samples common field values from marginal modes. The marginal-random baseline samples field values independently from empirical training marginals.

\subsection{Metric implementation}
\label{app:metric_implementation}

Scoring proceeds in two steps. \emph{First, pairing.} Predicted portfolios are greedily paired to ground-truth portfolios using indication similarity. For each predicted trial, the evaluator selects the best-matching unpaired ground-truth trial using a rule-based score over the indication and trial-description strings. Strings are lowercased, stripped of stage and severity prefixes, and normalized with a hand-curated oncology dictionary.

\emph{Second, field matching.} Once trials are paired, free-text fields (indication, trial description, comparator, and combination partners) are matched by cosine similarity between sentence embeddings (OpenAI \texttt{text-embedding-3-small}) with threshold $\tau{=}0.62$, while categorical and enumerated fields (phase, strategy, study type, allocation, masking, arm and enrollment buckets, region, and endpoint types) are matched exactly. Embedding matching credits paraphrases and abbreviations of the same disease (for example ``CLL'' and ``chronic lymphocytic leukemia'') that lexical string matching scores as misses, which otherwise understates the free-text agent outputs and overstates the gap to the models. The threshold $\tau{=}0.62$ is the midpoint between same-disease paraphrases (cosine $\approx 0.7$--$0.9$) and distinct diseases ($\approx 0.3$--$0.5$), and the method ranking is invariant for $\tau\in[0.50,0.75]$.

For paired trials, categorical and ordinal fields are evaluated by exact equality. Secondary endpoints are evaluated by set overlap. Combination partners and comparators are evaluated by a normalized substring rule. Field-match \% averages the percentage of matched fields per paired trial, and Strict F1 requires indication, phase, and strategy to all match within a paired prediction.

Because portfolio prediction is set-valued, evaluation depends on how predicted trials are paired to observed trials. We therefore treat Indication F1 as the headline alignment metric, use Field-match \% and Strict F1 to test increasingly structured agreement.

\subsection{Reproducibility and release}
\label{app:reproducibility}

The random seed is fixed for dataset splitting, reward-weighted oversampling, reward-model pair sampling, and fine-tuning. Each model directory contains the fine-tuning split, per-step \texttt{train\_history.jsonl}, and an MLflow run. Training runs on a single NVIDIA H100 80 GB. Per-model wall-clock time is approximately 13 to 30 minutes depending on split size. The full sweep contains 32 models from 8 splits and 4 training objectives.

\section{Supplementary results}
\label{app:supplementary_results}

\subsection{Random baselines: uniform, mode, and marginal}
\label{app:random}

We evaluated three random baselines. \textit{Uniform} samples each categorical dimension uniformly over the schema vocabulary. \textit{Mode} predicts the most-common training value for every dimension. \textit{Marginal} samples each dimension from the training marginal. All three predict a portfolio whose size is sampled from the training distribution of portfolio sizes, with mode-random using the rounded mean. Results are averaged over three seeds. Mode-random is strongest on every metric (Table~\ref{tab:app-random}) and is reported as ``Random'' in the main paper.

\begin{table}[H]
  \caption{Random baselines on the full held-out set ($n{=}432$).
  Mode-random predicts the same modal tuple per window; uniform and
  marginal sample per categorical dimension. Mode is the strongest of
  the three on every metric.}
  \label{tab:app-random}
  \centering
  \small
  \begin{tabular}{lrrr}
    \toprule
    Variant   & Ind.\ F1           & Field-match \%    & Strict F1         \\
    \midrule
    Uniform   & \phantom{0}7.3\%   & 2.9\%             & 0.1\%             \\
    \textbf{Mode}      & \textbf{11.4\%}    & \textbf{5.1\%}    & \textbf{0.6\%}    \\
    Marginal  & \phantom{0}6.6\%   & 3.3\%             & 0.2\%             \\
    \bottomrule
  \end{tabular}
\end{table}

Mode-random reaches 0.0\% Strict F1 on the 24-window post-August 2025 subset because Strict F1 requires the predicted trial to match on indication, phase, and strategy and to land on the right pair, and mode-random predicts the same fixed tuple in every window. On the full 432-window set the modal tuple coincides with the ground-truth tuple in roughly three windows, giving 0.6 percent. On the 24-window subset the intersection happened to be empty, which is a small-$n$ artefact rather than an evaluation error and is consistent with the 0.6 percent rate observed on the full set.

\subsection{Headline leaderboard with confidence intervals}
\label{app:headline}

Table~\ref{tab:headline-embed} reports the exact headline numbers visualised in Figure~\ref{fig:headline}: the contamination-controlled post-August 2025 cross-split subset ($n{=}24$). Brackets are 95\% confidence intervals from a nonparametric bootstrap over the held-out windows (percentile method). The intervals are wide at $n{=}24$, but on Strict F1 the reward-weighted model RW-BC excludes zero ($[1.7,29.2]$) while the base model and every agent are at or near zero.

\begin{table}[H]
  \caption{Contamination-clean leaderboard, post-August 2025 cross-split subset ($n{=}24$), with bootstrap 95\% CIs. Bold marks the best method per column. RW-BC is the reward-weighted model; the Qwen3-235B agent is shown as point estimates (CIs not computed).}
  \label{tab:headline-embed}
  \centering
  \small
  \setlength{\tabcolsep}{5pt}
  \begin{tabular}{@{}lccc@{}}
    \toprule
    Method                       & Indication F1 [95\% CI]   & Field-match \% [95\% CI]  & Strict F1 [95\% CI]       \\
    \midrule
    \textbf{RW-BC}               & \textbf{46.2} [30.0, 63.3]  & \textbf{56.4} [50.8, 61.9]  & \textbf{14.2} [1.7, 29.2]  \\
    LR-BC                        & 34.6 [19.4, 49.3]           & 55.6 [49.2, 62.2]           & \phantom{0}2.8 [0.0, 8.3]   \\
    BC                           & 36.1 [20.8, 52.5]           & 48.9 [39.9, 57.1]           & \phantom{0}8.3 [0.0, 20.8]  \\
    IQL                          & 32.1 [17.5, 48.2]           & 53.5 [47.4, 59.5]           & \phantom{0}6.2 [0.0, 16.7]  \\
    Qwen-2.5-7B             & 33.5 [23.9, 43.4]           & 34.0 [29.6, 39.0]           & \phantom{0}0.0 [0.0, 0.0]   \\
    \cmidrule(lr){1-4}
    Qwen3-235B agent             & 25.0                        & 35.6                        & \phantom{0}0.0              \\
    GPT-5.4 agent                & 22.2 [9.9, 35.1]            & 28.7 [21.3, 35.5]           & \phantom{0}2.1 [0.0, 6.2]   \\
    Gemini 3.1 Pro agent         & 21.5 [9.0, 35.4]            & 29.0 [20.8, 37.2]           & \phantom{0}0.0 [0.0, 0.0]   \\
    Opus 4.5 agent               & 11.4 [4.4, 19.3]            & 38.3 [33.5, 43.3]           & \phantom{0}0.0 [0.0, 0.0]   \\
    \bottomrule
  \end{tabular}
\end{table}

\subsection{Reward component ablation and credit assignment}
\label{app:reward-ablation}

The per-trial reward (Eq.~\ref{eq:reward}) attributes approval and failure to the specific trial that earned them rather than pooling them over a window. At the canonical first-line NSCLC fork, KEYNOTE-024 (the PD-L1-selected trial that succeeded) scores $+11.0$, a pivotal-approval credit plus attributed revenue, while CheckMate-026 (the all-comers trial that failed) scores $-1.0$: a clean separation assigned directly to the two decisions.

We decompose the outcome reward across its terms. Table~\ref{tab:app-reward-components} reports each term's mean contribution over the 881 windows, its correlation with the total reward, and a leave-one-out measure of how much dropping it perturbs the training weights. The reward is dominated by approval-linkage (correlation $0.98$ with the total); the fixed revenue and failure coefficients move the weights little (dropping the failure penalty shifts the mean absolute weight by only $0.095$). The reward's signal is therefore an approval-linkage signal, which is a program-level quantity, consistent with the task-hardness probe (state$\to$reward $R^2\!=\!0.49$ versus decision$\to$reward $R^2\!=\!0.025$). This is why we do not over-claim a finely tuned reward: the reward's strength comes from grounding it in real approval outcomes.

\begin{table}[H]
  \caption{Reward component leave-one-out over the 881 saved window rewards. ``corr'' is the correlation of each term with the total reward; the last column is the mean absolute change in training weight when the term is dropped. Approval-linkage dominates.}
  \label{tab:app-reward-components}
  \centering
  \small
  \begin{tabular}{lrrr}
    \toprule
    Component              & mean   & corr with total   & $\overline{|\Delta w|}$ on drop \\
    \midrule
    approval reward        & $+2.87$ & \textbf{$+0.98$}  & 0.358 \\
    revenue reward         & $+0.63$ & $+0.21$           & 0.215 \\
    failure penalty        & $+0.18$ & $-0.24$           & 0.095 \\
    \bottomrule
  \end{tabular}
\end{table}

The IQL model is robust to its hyperparameters: the per-example weights it induces are rank-correlated at $\geq0.81$ across the discount $\gamma$ and expectile $\tau$ we swept, and the inverse-temperature $\beta$ is essentially inert over the same range. Together with the component ablation above, this means neither the reward terms nor the value-model settings need fine tuning to obtain the reported behavior.

Reward-weighted cloning with the specified reward is the strongest objective overall---expected when the extra method complexity of a learned reward or of value learning is not repaid by the data scale. Learned-reward weighting fits a regression whose target is the specified reward itself, so its ceiling is reward-weighted cloning and the approximation only adds variance. Value-based IQL must learn $Q$/$V$ across delayed, sparse rewards from few episodes, producing high-variance advantages that down-weight windows toward the clipping floor and lose indication coverage; reward-weighted cloning instead uses direct, Monte-Carlo-style outcome attribution with lower variance at small data. This is consistent with reports that behavioral cloning can match or beat offline value-based RL when data are limited \citep{kumar2022when,emmons2022rvs,brandfonbrener2022when,wang2020critic}.

\subsection{Per-dimension paired F1}
\label{app:per-dim}

Table~\ref{tab:app-perdim} reports per-dimension F1 for each structured trial field on the post-August 2025 subset ($n{=}24$).

\begin{table*}[h]
  \caption{Per-dimension F1 on the post-August 2025 subset ($n{=}24$). Bold marks the best system per field. No single system dominates: the offline models lead on indication, phase, masking, comparator, and primary endpoint, while the Gemini 3.1 Pro agent is competitive or best on several categorical design fields.}
  \label{tab:app-perdim}
  \centering
  \scriptsize
  \setlength{\tabcolsep}{4pt}
  \begin{tabular}{@{}l rrrr r rrrr@{}}
    \toprule
    Field                    & IQL             & RW-BC           & LR-BC           & BC              & Qwen-2.5-7B            & GPT-5.4         & Gemini 3.1 Pro  & Opus 4.5        & Qwen3-235B      \\
    \midrule
    indication               & 38.7            & \textbf{58.6}   & 51.9            & 41.9            & 53.8            & 40.0            & 46.7            & 32.5            & 41.0            \\
    trial description        & 29.0            & 34.5            & 40.7            & 38.7            & 30.8            & \textbf{42.9}   & 33.3            & 40.0            & 38.5            \\
    phase                    & 38.7            & \textbf{51.7}   & 48.1            & 32.3            & 15.4            & \phantom{0}2.9  & 13.3            & 27.5            & \phantom{0}7.7  \\
    strategy                 & 25.8            & 24.1            & 25.9            & \textbf{32.3}   & 20.5            & 22.9            & 23.3            & 20.0            & 23.1            \\
    study type               & 77.4            & \textbf{86.2}   & 77.8            & 83.9            & 74.4            & 60.0            & 80.0            & 85.0            & 71.8            \\
    allocation               & 48.4            & 55.2            & \textbf{63.0}   & 61.3            & 35.9            & 42.9            & 56.7            & 42.5            & 30.8            \\
    masking                  & 87.1            & 86.2            & 92.6            & 87.1            & 74.4            & 77.1            & \textbf{100.0}  & 70.0            & 87.2            \\
    n\_arms                  & \textbf{64.5}   & 37.9            & 37.0            & 45.2            & 12.8            & 22.9            & 26.7            & \phantom{0}5.0  & 17.9            \\
    enrollment               & 32.3            & \textbf{44.8}   & 44.4            & 41.9            & 25.6            & 37.1            & 23.3            & 27.5            & 25.6            \\
    region                   & \textbf{90.3}   & 89.7            & 77.8            & 58.1            & 48.7            & 37.1            & 60.0            & 47.5            & 51.3            \\
    primary endpoint         & 35.5            & 41.4            & 44.4            & \textbf{54.8}   & 15.4            & 22.9            & 10.0            & 22.5            & 12.8            \\
    comparator               & 67.7            & 72.4            & \textbf{85.2}   & 74.2            & 10.3            & 45.7            & 20.0            & 37.5            & 15.4            \\
    combo partners           & 45.2            & 37.9            & 44.4            & 32.3            & 28.2            & 34.3            & \textbf{46.7}   & 25.0            & 35.9            \\
    \bottomrule
  \end{tabular}
\end{table*}

No single system dominates every field, and leadership is mixed across the fine-tuned models. RW-BC leads on indication, phase, study type, and enrollment; LR-BC on allocation and comparator; BC on strategy and primary endpoint; IQL on the multi-arm and region fields. The frontier agents are competitive only on a few categorical fields (the Gemini 3.1 Pro agent ties masking and leads the free-text trial description). This is consistent with the headline story: the agents can often get individual design choices right, but only the offline models assemble them into the correct whole decision, which is why the agents trail sharply on Strict F1 despite matching individual dimensions here.

\subsection{Design accuracy conditional on indication}
\label{app:conditional}

To isolate ``when a model proposes the right trial, how well does it get the design?'' we score per-field accuracy on indication-matched pairs only, without a volume penalty, on the contamination-controlled temporal split. Table~\ref{tab:app-conditional} reports phase accuracy and mean design-field accuracy. Conditional on a correct indication, the offline models predict the design substantially better than the base model (LR-BC leads design at 53.9\% versus 33.5\% for the base; BC leads phase at 56.2\%). The GPT-5.4 agent's design is competent when it identifies the right trial (44.8\%), but it matches the correct indication far less often, so its deficit is decision targeting, not design knowledge. Cell counts are small (27 to 41 matched pairs) and absolute design accuracy plateaus near 50\% even for the best model, consistent with the associational ceiling.

\begin{table}[H]
  \caption{Per-field accuracy conditional on an indication match, on the post-September 2024 temporal split. DESIGN is the mean over the design fields; matched pairs is the conditioning sample size.}
  \label{tab:app-conditional}
  \centering
  \small
  \begin{tabular}{@{}lrrr@{}}
    \toprule
    Arm              & PHASE  & DESIGN & matched pairs \\
    \midrule
    BC               & \textbf{56.2\%} & 52.3\% & 32 \\
    LR-BC            & 44.8\% & \textbf{53.9\%} & 29 \\
    IQL              & 47.1\% & 50.5\% & 34 \\
    RW-BC            & 35.3\% & 48.4\% & 34 \\
    GPT-5.4 agent    & 22.2\% & 44.8\% & 27 \\
    Qwen-2.5-7B      & 19.5\% & 33.5\% & 41 \\
    \bottomrule
  \end{tabular}
\end{table}

\subsection{Precision and recall decomposition of Indication F1}
\label{app:precision-recall}

The headline tables report Indication F1, the harmonic mean of precision and recall over indication matches. Decomposing it, alongside mean predicted and ground-truth portfolio sizes, exposes a behavioral difference the F1 number alone hides (Table~\ref{tab:app-pr}).

\begin{table}[H]
  \caption{Indication-level precision, recall, and F1 on the post-August 2025 subset ($n{=}24$), with the mean predicted ($\bar{n}_\mathrm{pred}$) and ground-truth ($\bar{n}_\mathrm{gt}$) portfolio sizes per window. Bold marks the best method per column.}
  \label{tab:app-pr}
  \centering
  \small
  \begin{tabular}{lrrrrr}
    \toprule
    Method            & Precision        & Recall           & F1               & $\bar{n}_\mathrm{pred}$ & $\bar{n}_\mathrm{gt}$ \\
    \midrule
    Qwen-2.5-7B  & 26.0\%           & \textbf{52.1\%}  & 33.5\%           & 3.6                     & 1.7                   \\
    Qwen3-235B agent  & 22.5\%           & 29.9\%           & 25.0\%           & 2.9                     & 1.7                   \\
    GPT-5.4 agent     & 19.1\%           & 28.1\%           & 22.2\%           & 3.3                     & 1.7                   \\
    Gemini 3.1 Pro agent  & 21.7\%       & 22.9\%           & 21.5\%           & 2.1                     & 1.7                   \\
    Opus 4.5 agent    & \phantom{0}7.7\% & 24.0\%           & 11.4\%           & 7.0                     & 1.7                   \\
    \midrule
    BC                & 38.2\%           & 37.5\%           & 36.1\%           & 2.4                     & 1.7                   \\
    \textbf{RW-BC}    & \textbf{52.9\%}  & 46.9\%           & \textbf{46.2\%}  & 1.6                     & 1.7                   \\
    LR-BC             & 39.6\%           & 37.5\%           & 34.6\%           & 1.8                     & 1.7                   \\
    IQL               & 35.8\%           & 35.8\%           & 32.1\%           & 2.5                     & 1.7                   \\
    \bottomrule
  \end{tabular}
\end{table}

Two patterns are visible in Table~\ref{tab:app-pr}. The tool agents over-predict: the Opus 4.5 agent emits 7.0 trials per window and GPT-5.4 3.3, against roughly 1.7 in the ground truth, which pads recall but dilutes precision into the 8 to 22 percent range. The offline-trained models are calibrated in the opposite direction: RW-BC predicts only 1.6 trials per window yet reaches the best precision (52.9 percent) and F1, and the other offline models track the sponsor portfolio size closely. This decomposition explains why the agents look superficially competitive on indication overlap but collapse on Strict F1: a verbose portfolio rarely matches the full indication, phase, and strategy tuple on any single trial even when one of its many predictions hits the right disease.

\subsection{Indication F1 versus Strict F1}
\label{app:pareto}

\begin{figure}[H]
  \centering
  \includegraphics[width=0.6\columnwidth]{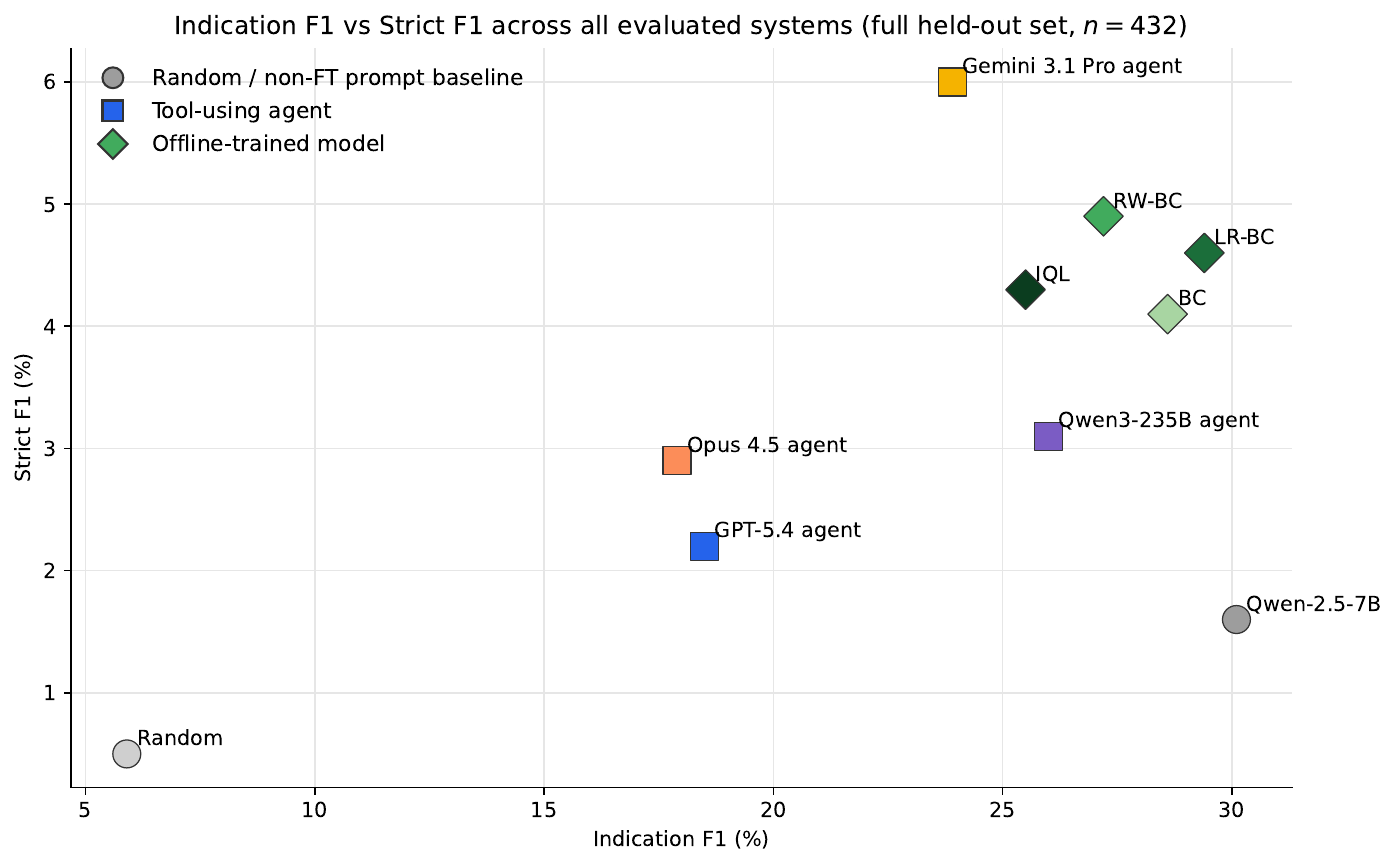}
  \caption{Indication F1 versus Strict F1 across methods on the full held-out set ($n{=}432$). Higher and more rightward is better. The offline-trained models (diamonds) occupy the upper-right, scoring well on both axes; the tool-using agents (squares) sit lower; the non-fine-tuned Qwen-2.5-7B reaches high Indication F1 but near-zero Strict F1 (it names the right disease without the matching structured trial), and the random baseline sits near the origin.}
  \label{fig:app-pareto}
\end{figure}

Plotting Indication F1 against Strict F1 separates methods that get the disease right from methods that get the whole structured trial right. Figure~\ref{fig:app-pareto} shows the methods on these two axes for the full held-out set.

On the full set the offline-trained models occupy the upper-right, scoring well on both axes. The non-fine-tuned Qwen-2.5-7B reaches comparable Indication F1 (30.1) but near-zero Strict F1 (1.6), naming the right disease without the matching structured trial, while the offline models convert more of their indication hits into full-design matches. The tool-using agents trail, though the Gemini agent is competitive on Strict.

\subsection{Reduced prompt}
\label{app:no-tools-agent}

To separate the effect of the prompt format from the information the prompt carries, we compare our structured prompt against a bare agent prompt with an empty tool list. Table~\ref{tab:app-no-tools} reports Indication F1 on GPT-5.4 and Qwen-2.5-7B, the two backbones for which we have complete coverage. The structured prompt is stronger for both backbones on both subsets. On the full held-out set by roughly five points for GPT-5.4 and seven for Qwen, and on the post-August 2025 contamination-controlled cell by ten and fourteen points respectively. This demonstrates that the context and information supplied to the model can improve performance.

\begin{table}[H]
  \caption{Indication F1 under the structured prompt versus a bare agent prompt (the agent system prompt with an empty tool list). Neither variant receives retrieved evidence.}
  \label{tab:app-no-tools}
  \centering
  \small
  \begin{tabular}{llrrr}
    \toprule
    Backbone     & Subset                       & Prompt     & Agent prompt      & $\Delta$ \\
    \midrule
    GPT-5.4      & Full set ($n{=}432$)         & 19.9\%           & 15.0\%            & $-4.9$ \\
    GPT-5.4      & Post-cutoff ($n{=}24$)       & 22.0\%           & 11.4\%            & $-10.6$ \\
    \midrule
    Qwen-2.5-7B  & Full set ($n{=}432$)         & 30.4\%           & 23.2\%            & $-7.2$ \\
    Qwen-2.5-7B  & Post-cutoff ($n{=}24$)       & 32.0\%           & 17.6\%            & $-14.4$ \\
    \bottomrule
  \end{tabular}
\end{table}

\end{document}